\documentclass[a4paper,final]{article}

\usepackage{graphicx}
\usepackage{amsmath}
\usepackage{graphics,epsfig,color,times}
\graphicspath{{graphics/}}
\usepackage{amsmath}
\usepackage{amssymb}
\usepackage{xspace}
\usepackage{enumitem}

\usepackage{hyperref}
\usepackage{float}

\usepackage[top=2.5cm, bottom=2.5cm, left=3cm, right=3cm]{geometry}
\usepackage{float}

\newcommand{\T}{\ensuremath{\top}}
\newcommand{\ie}{i.\,e.~}
\newcommand{\eg}{e.\,g.~}
\newcommand{\wrt}{w.\,r.\,t.~}
\newcommand{\g}[1]{
  \ifthenelse{\equal{#1}{(}}{ \left( }{ \ifthenelse{\equal{#1}{)}}{ \right)}%
    { \ifthenelse{\equal{#1}{[}}{ \left[}{ \ifthenelse{\equal{#1}{]}}{ \right]}{#1}}}}}

\newcommand{\fig}[1]{Fig.~\ref{#1}}
\newcommand{\figapp}[1]{Supplementary Fig.~\ref{#1}}

\newcommand{\eqn}[1]{Eq.~\eqref{#1}} 
\newcommand{\eqnp}[1]{(\ref{#1})} 
\newcommand{\eqns}[2]{Eqs.~(\ref{#1}, \ref{#2})} 
\renewcommand{\sec}[1]{Sect.~\ref{#1}}
\newcommand{\app}[1]{Supplementary Section \ref{#1}} 
\newcommand{\vid}[1]{Video~\ref{vid:#1}}

\title{A novel plasticity rule can explain the development of sensorimotor intelligence}
\author{Ralf Der$^1$ and Georg Martius$^{1,2}$\\[.5em]
{\normalsize $^1$\,Max Planck Institute for Mathematics in the Science, Leipzig, Germany}\\
{\normalsize $^2$\,IST Austria, Klosterneuburg, Austria}}

\begin{document}
\maketitle

\bibliographystyle{naturemag2}

\begin{abstract}
Grounding autonomous behavior in the nervous system is a fundamental challenge for neuroscience.
In particular, the self-organized behavioral development provides more questions than answers.
Are there special functional units for curiosity, motivation, and creativity?
This paper argues that
these features can be grounded in synaptic plasticity itself,
without requiring any higher level constructs.
We propose differential extrinsic plasticity (DEP) as a new synaptic rule for self-learning systems
and apply it to a number of complex robotic systems as a test case.
Without specifying any purpose or goal, seemingly purposeful and adaptive behavior is developed,
 displaying a certain level of sensorimotor intelligence.
These surprising results require no system specific modifications of the DEP rule
 but arise rather from the underlying mechanism of spontaneous symmetry breaking
due to the tight brain-body-environment coupling.
The new synaptic rule is biologically plausible and it would be an interesting target
 for a neurobiolocal investigation.
We also argue that this neuronal mechanism may have been a catalyst in natural evolution.
\end{abstract}





\section{Introduction}
Research in neuroscience produces understanding of the brain on many different levels.
At the smallest scale there is enormous progress in understanding mechanisms of neural signal transmission and processing~\cite{SquireEtAl2013:FundamentalNS,Spitzer2012:neurotransmitter,Scheiffele2003:CellSignaling,Rourke2012:synapsis}.
At the other end, neuroimaging and related techniques enable the creation of a global understanding of the
brain's functional organization~\cite{PowerEtAl2011:FuncBrainNet,AllanBrainAtlas:2014}.
However, there remains a gap in binding these results together, leaving open the question of
how all these complex mechanisms interact~\cite{Grillner2005:IntegrativeNS, AlivisatosEtAl2012:BAM,KandelEtAl2013:BigNSProjects}.
This paper advocates for the role of self-organization in bridging this gap.
We focus on the functionality of neural circuits
acquired during individual development by processes of self-organization --- making
complex global behavior emerge from simple local rules.

Donald Hebb's formula ``cells that fire together wire together''~\cite{Hebb1949} may be seen as an early example of
such a simple local rule which has proven successful in building associative memories and perceptual functions~\cite{Hopfield1982, Keysers2004:Hebbian}.
However, Hebb's law and its successors like BCM~\cite{BCM1982} and STDP~\cite{Markram1997:STDP,Gerstner1996:STDPModel} are restricted to   scenarios
 where the learning is driven passively by an externally generated data stream.
However, from the perspective of an autonomous agent,
sensory input is mainly determined by its own actions.
The challenge of behavioral self-organization
requires a new kind of learning that
bootstraps novel behavior out of the self-generated past experiences.

This paper introduces a rule which may be expressed as ``chaining together what changes together.''
This rule both takes into account temporal structure
and establishes the contact to the external world by directly relating the  behavioral level to the synaptic dynamics.
These features together provide a mechanism for bootstrapping behavioral patterns from scratch.

This synaptic mechanism is neurobiologically plausible and
naturally raises the question of whether it is present in living beings.
This paper aims to encourage such  initiatives by
using bioinspired robots as a methodological tool.
Admittedly, there is a large gap between biological beings and such robots.
Yet, in the last decade, robotics has seen a change of paradigm
from classical AI thinking to embodied AI~\cite{PfeiferBongar2006:BodyShapesThink,Pfeifer2007}
which recognizes the role of embedding the specific body in its environment.
This has moved robotics closer to biological systems and supports their use as a testbed for neuroscientific hypotheses~\cite{Floreano2014:RoboticsAndNeuroscience}.

We deepen this argument by presenting  concrete results showing that the proposed synaptic plasticity rule
generates a large number of phenomena which are 
important for neuroscience. We show that, up to the level of sensorimotor
contingencies, self-determined behavioral development can be grounded
in synaptic dynamics, without having to postulate higher level constructs
such as intrinsic motivation, curiosity, or a specific reward system.
This is achieved with a very simple neuronal control structure by outsourcing much of the complexity to the embodiment (the idea of morphological computation~\cite{Pfeifer2009,Hauser2011}).

The paper comes with a supporting material containing video clips and technical detail
that is available at \href{http://playfulmachines.com/DEP}{\small\texttt{http://playfulmachines.com/DEP}}.
We recommend to start with \vid{DEP-demo} providing a brief overview.

\section{Grounding behavior in the synaptic plasticity}

We consider generic robotic systems,
a humanoid and a hexapod robot,
in physically realistic simulations using \textsc{LpzRobots}~\cite{lpzrobots10}. 
These robots are mechanical systems of rigid body primitives
linked by joints. With each joint $i$, there is an angular  motor for realizing the new joint angles
$y_i$ as proposed by the controller network
and there is a sensor measuring the true joint angle $x_i$ (like muscle spindles).
The implementation of the motors is similar to muscle/tendon driven systems
 by being compliant to external forces.

\subsection{Controller network}

One theme of our work is structural simplicity, building on the paradigm that complex behavior may
emerge from the interaction of a simple neural circuitry with the complex external world.
Specifically, the controller is a network of rate-coded neurons transforming sensor values $x = (x_1,x_2,\dots,x_n)$ into motor commands
$y=(y_1,y_2,\dots,y_m)$.
In the application, a one-layer feed-forward network is used, described as
\begin{equation}
  y_i= g\left(\sum_{j=1}^n C_{ij}x_{j}+h_i\right)
  \label{eqn:DHL10}
\end{equation}
for neuron $i$, where
$h_i$ is the threshold. We use $\tanh$-neurons, \ie the activation function $g(z)=\tanh(z)$ to get motor commands between +1 and -1. This type of neurons is chosen for simplicity,
 but our approach can be translated into a neurobiologically more realistic setting.
The setup is displayed in \fig{fig:controller}.

This controller network may appear utterly oversimplified.
Commonly, and in particular in classical Artificial Intelligence,
a certain behavior is seen as the execution of a plan devised by the brain.
This would require a highly organized internal brain dynamics,
which could never be realized by the simple one-layer network.
However, in this paper behavior is an emerging mode in the dynamical system formed by brain, body, and environment~\cite{PfeiferBongar2006:BodyShapesThink}.
As we demonstrate here, by the new synaptic rule, the above simple feed-forward network can generate
a large variety of motion patterns in complex dynamical systems.

\subsection{Chaining together what changes together}
\label{sec:DHLrule}
\begin{figure}
  \centering
  \includegraphics[width=.8\linewidth]{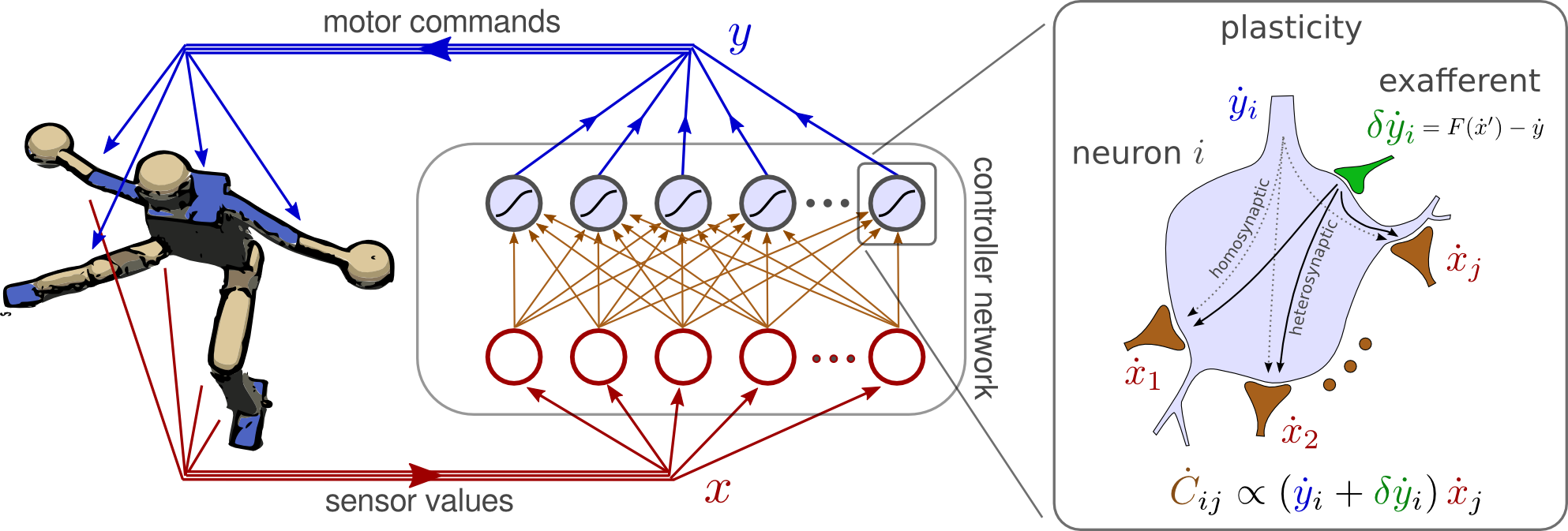}%
  \caption{{\bf Controller network connected to the humanoid robot.}
    The proposed differential extrinsic plasticity (DEP) rule is
    illustrated in the insets on the right.
    In addition to the homosynaptic term $\dot y$ of plain differential Hebbian learning,
    the DEP \eqnp{eqn:DHL100simple} has the exafferent signal $\delta \dot y$
    that may be integrated by a heterosynaptic mechanism.
    In the simplest case, \eg with the humanoid robot,
    the inverse model $F$ is a one-to-one mapping of sensor to motor values.
  }
  \label{fig:controller}
\end{figure}

When learning the controller \eqnp{eqn:DHL10} with a Hebbian law, the rate of change $\dot C_{ij}$ of synapse $C_{ij}$ would be proportional
to the input $x_j$ into the synapse of neuron $i$ multiplied by its activation  $y_i$, \ie
$\dot C_{ij}\propto y_i x_j$.
However, in the concrete settings,  this rule produces typically fixed-point behaviors.
It was suggested earlier~\cite{kosko1986differential,roberts1999DHL,lowe2011DHL} that 
time can come into play in a more fundamental way if the so called differential Hebbian learning (DHL) is used, \ie replacing the neuronal activities by their rates of change,
so that  $\dot C_{ij}\propto \dot y_i \dot x_j$ (the derivative of $x$ \wrt time is denoted as $\dot x$).
This rule focuses on the dynamics as there is only a change in behavior if the system is active.
As demonstrated in \sec{sec:methods}, this may produce interesting behaviors
 but in general it lacks the drive for exploration that is vital for a developing system.

The main reason for the lack in behavioral richness is seen in the product structure of both learning rules
which involves the motor commands $y$ generated by the neurons themselves.
Trivially, once $y=0$, learning and any change in behavior is stopping altogether.
Now, the idea is to lift this correlative structure entirely to the level of the outside world,
enriching learning by the reactions of the physical system to the
controls.

Let us assume the robot has a basic understanding of the causal
 relations between actions and sensor values.
In our approach, this is realized by an inverse model which approximately relates the
current sensor values $x'$ back to its causes, the motor commands $y$ having a certain time lag \wrt $x'$. As the model is  never  exact,
it will reconstruct (the efference copy) $y$ with a certain mismatch $\delta y$.
Formulated in terms of the rates of change, we write
\begin{equation}
\dot y + \delta \dot y = F\left(\dot x'\right)
\label{eqn:tildey}
\end{equation}
with $F$ the model function and
 $\delta y$ containing all (physical) effects that cannot be captured by the model.

The aim of our approach is to make the system sensitive to these effects.
This is achieved by replacing $\dot y$ of the DHL rule with $\tilde{\dot{y}}  = \dot y + \delta \dot y$, so that
\begin{equation}
  \tau\dot C_{ij}= \tilde{\dot{y_i}} \dot x_j -  C_{ij}
    \label{eqn:DHL100simple}\end{equation}
where $\tau$ is the time scale for this synaptic dynamics and $-C_{ij}$ is a
damping term, see \fig{fig:controller}. Because of the normalization introduced below, we do not
 need an additional scaling factor for the decay time.
In principle, the model is relating the changes in sensor values caused by the robot's behavior
 back to the controller output and the learning rule is extending this chain further down to the synaptic weights.
This is the decisive step in the ``chaining together what changes together'' paradigm.
The $\delta \dot y$ in $\tilde{\dot{y}}$ contains all physical effects that are extrinsic
to the system as they are not captured by the model.
They are decisive for exploring the behavioral capabilities of the system.
That is why we call the new mechanism defined by \eqn{eqn:DHL100simple} \emph{differential extrinsic plasticity} (DEP).

Optionally, the threshold terms $h_i$ \eqnp{eqn:DHL10} can also be given a dynamics which we simply define as
\begin{equation}
\tau_h  \dot h_i =-y_i
  \label{eqn:biasdyn}
\end{equation}
where $\tau_h$ defines an empirical time scale. The idea is to drive
the neurons away from their saturation regions (close to $y=\pm 1$).
As the experiments will demonstrate, using the threshold dynamics favors periodic motion patterns.

The realization of inverse models is a notoriously difficult task if the model is to capture as much as possible
 of the system dynamics.
Here, the role of the model is different, as
the extrinsic term $\delta \dot y$ takes the lead in the unfolding of motion patterns.
Thus, the task of the model is just to separate between intrinsic and extrinsic effects.
Yet, it cannot be chosen arbitrarily as it has to reflect the basic causal relations between sensor and motor values.
Depending on the capabilities of the model different routes in the development of behaviors are taken.
Throughout this and many
other work~\cite{DerMartius11}, a simple linear model was appropriate, stipulating $F\left(\dot x'\right)=M\dot x'$ where $M$ is a weight matrix,
relating the sensor values back to motor commands, \ie
 \begin{equation}
\tilde{\dot{y}}_i= \sum_j M_{ij}\dot x_j'\ .\label{eqn:hypotheticalcommand}
\end{equation}
The weights $M_{ij}$ of the inverse model can be learned off-line from $(\dot x',\dot y)$ pairs in some idealized situations, like the robot being free of any external forces and constraints, using classical methods (low frequency motor babbling).
Alternatively, it can be set by hand in order to guide the self-organization process as done in the experiments with the Hexapod below. 

In order to make the whole system exploratory and curious for new behaviors, we introduce an
appropriate normalization of the synaptic weights $C$ and an empirical \emph{gain} factor $\kappa\sim 1$.
The latter regulates the overall feedback strength in
the sensorimotor loop. If chosen in the right range
 the extrinsic perturbations get amplified and active behavior can be maintained,
see \sec{sec:methods} for details.
This specific regime, called the edge of chaos, is argued to be a vital characteristic of life and development~\cite{Langton1990:edge,Kauffman1995:edge,bertschinger2004:edgeofchaos}. 
The normalization is supported by neurophysiological findings on synaptic normalization~\cite{Carandini2011:Normalization}
 such as homeostatic synaptic plasticity~\cite{Turrigiano2004},
 and the balanced state hypothesis~\cite{Tsodyks1995:balancedstate,MonteforteWolf2010:balancedstate_entropyproduction}.

While $\kappa$ is seen to regulate the overall activity, $\tau$ is found to regulate the
degree of exploration.
As described in \sec{sec:suppl:How}
the system realizes a ``search and converge'' strategy,
wandering between metastable attractors (such as walking patterns) with possibly very long transients.
The time spent in an attractor (a certain motion pattern) is regulated by $\tau$.
At the behavioral level, this is reflected by the emergence of a great variety of spatio-temporal patterns---the
global order obtained from the simple local rules given by \eqns{eqn:DHL100simple}{eqn:biasdyn}.

\subsection{Behavior as broken symmetries}\label{sec:BasSSB}

In order to understand how the very specific behaviors can emerge from the generic synaptic mechanism,
 we have to consider the role of the symmetries.
For a discussion, let  us consider the system in
what we call its least biased initialization, \ie  putting $C_{ij}=0$ and $h_i=0$ so
that all actuators are at their central position.
In this situation, the agent obeys a maximum number of symmetries. These are the obvious geometric symmetries but
also a bunch of dynamical ones originating from the invariance of the physical system against
certain transformations, like inverting the sign of a joint angle.
Technically, the symmetries are seen directly by a linear expansion of the system around the resting situation.
As the learning rule does not introduce any symmetry breaking preferences, motion can set in only by a spontaneous breaking
of the symmetries.
In this picture, behavior corresponds to broken symmetry  (in space and time) and development to a sequence of spontaneous symmetry breaking  events.
This is the very reason for the rich phenomenology observed in the
experiments, explaining the emerging dimensionality reduction which makes the approach scalable.

Self-organizing behavior as a result of symmetry breaking was observed
before~\cite{DerGSO2012,DerMartius13} with precursors of the newly proposed learning rule.
%
More specifically, related unsupervised learning rules based on
 the principle of homeokinesis~\cite{DerMartius11} and the maximization of
 predictive information~\cite{MartiusDerAy2013,ay2012information,ay08:predinf_explore_behavior,ZahediAyDer2010:HigherCoordination} were studied,
 which however differ in being biologically implausible due to matrix inversions and
 leading to less specific behavioral modes.

\subsection{Neurobiological implementation}

In order to understand how the DEP rule can be implemented neurobiologically,
we note first that  $x'$ is just the reafference caused by $y$.
Commonly,  the contribution in $x'$ that cannot be accounted for by the (forward) model is called the exafference.
Our extrinsic learning signal $\delta y$ is the preimage of this exafference using the forward model.
With the inverse models used here, this preimage can be obtained explicitely by a simple neural circuitry
 calculating the difference between the output of the inverse model $F(x')$ and the efference copy of $y$, see \eqn{eqn:tildey}.
By feeding this signal back to the neuron by an additional synapse,
the output of the neuron is shifted from $y$ to $\tilde y$.
With the modified output,  the new synaptic rule corresponds to classical (differential) Hebbian learning.
This procedure, although pointing a way to a concrete neurobiological implementation, is a little awkward
as the additional signal has to be subtracted again from the neuron output before sending the latter to the motors.

Another possibility is the inclusion of the extrinsic term by a heterosynaptic or extrinsic plasticity~\cite{Bailey2000:heterosynaptic}
mechanism as illustrated in \fig{fig:controller}.
The additional input from $\delta y$ has to simulate the effect of depolarization (firing)
 for the otherwise unchanged synaptic plasticity. This may
  by accomplished by G protein~\cite{Neves2002:GProtein} signaling or the enhanced/inhibited expression of synapse-associated proteins~\cite{Rumbaugh2003:SAP97}, or via other intra-cellular mechanism.

\section{Results}\label{sec:results}

By the following series of experiments, we demonstrate
the potential of the new synaptic plasticity for the self-organization of behavior.
Interestingly, the emerging behaviors seem to be purposeful,
 as if the learning system develops solutions for different tasks
 like locomotion, turning a wheel, and so forth.
In order not to plug in such a task orientation, we always use the same neural network (with appropriate number of motor
neurons and sensor inputs) with the DEP rule of \eqn{eqn:DHL100simple}, and start all
experiments in its  least biased initialization. 

\subsection{Early individual development}
In a first set of experiments, we
study the very early stage of individual development when sensorimotor contingencies are being acquired.
The common assumption is that sensorimotor coordination is developed
by learning to ``understand'' sensor responses caused by spontaneous muscle contractions (called motor babbling in robotics).
However, this fails by the abundance of sensorimotor contingencies as may be demonstrated by
a coarse assessment for our humanoid robot.
If we postulate that each of the $m$ motor neurons has only 5 different output values (rates)
we have in each step $5^m$ possible choices. With $m=18$ for the
humanoid robot and 50 steps per second,
 a motion primitive of $1$\,sec duration has $5^{18^{50}}$ realizations.
The number of possible sensor responses
is of the same dimension so that there is no way of probing and storing all sensorimotor contingencies.
Alternatively, realizing search by randomly choosing the synaptic connections
($C_{ij}$ with normalization, gain, and threshold dynamics)
yields another tremendous number of possible behaviors, even if we restrict
ourselves to the simplified nervous system formulated in \eqn{eqn:DHL10}.

On the contrary, there is no randomness involved in the DEP approach.
Both the physical dynamics and the plasticity rule
are purely deterministic.
Nevertheless, at the behavioral level, the above mentioned  ``search and converge'' strategy,
creates a large variety of highly
active, but time-coherent motion patterns, depending on
the initial kick,  the combination of the parameters $\tau$ and $\kappa$, and the
body-environment coupling.

\paragraph{Self-organized crawling behavior:}

\begin{figure}
  \includegraphics[width=\linewidth]{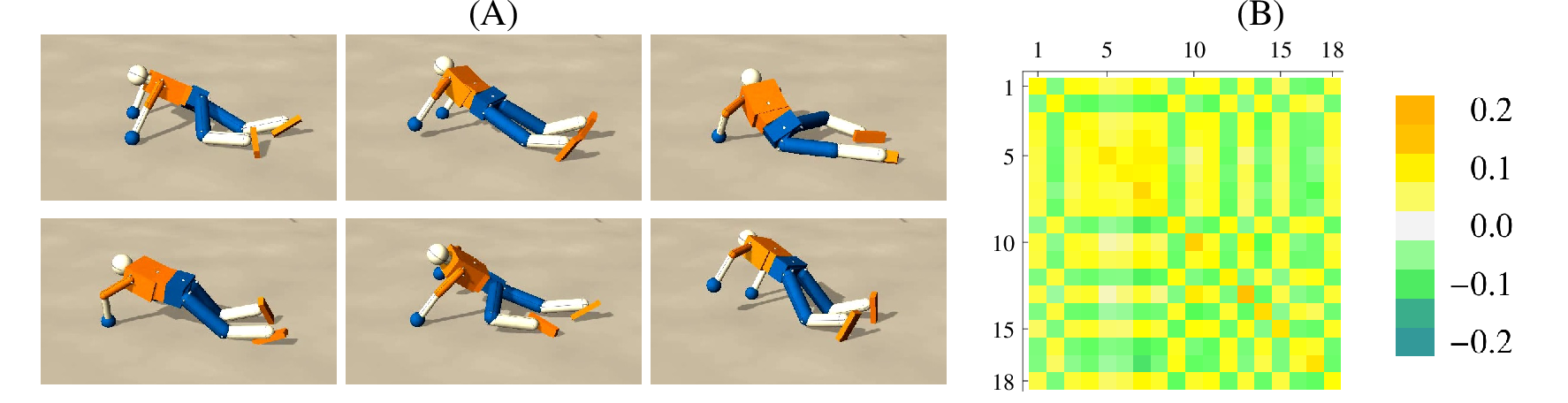}
  \caption{{\bf Behavior exploration of the humanoid robot: crawling like motion patterns when on the floor} (A).
   (B) shows the corresponding controller matrices $C$ revealing a definite structure.
   Note that the threshold dynamics was included here which is an important factor for this highly organized behavior.
   \label{fig:motorbabbling}
  }
\end{figure}

In the following example we consider the humanoid robot on level ground with a certain friction and elasticity.
Additionally to DEP, in this case, the bias dynamics \eqnp{eqn:biasdyn} was used which supports oscillatory behaviors.
The robot starts upright with all joints in their center position ($y_i=0$) slightly above the ground
 and falls down to its feet and then to its arms, such that the robot is laying face down.
The first contact with the ground and the gravitation
 exert forces on the joints that lead to non-zero sensor readings.
This creates a first learning signal and leads to small movements which get more and more amplified and shaped
 by the body-environment interaction.
As a result we observe more and more coordinated movements such as the swaying of the
hip to either side.
A long transient of different behavioral patterns follows ending up in a
 forward crawling mode, see \fig{fig:motorbabbling} and \vid{humcrawl}.
This mode is meta-stable and can be left by perturbations or changes in the parameters.
Interestingly, mainly forward locomotion is emerging which is due to the specific geometry
 of the body.
For an external observer this looks as if the robot is following a specific purpose, exploring its environment,
 which is not built in but emerges.
When the parameters of the body are changed, \eg the strength of certain actuators,
 different behaviors will come out. For instance,  a low crawling mode is generated if the arms are weaker.

\subsection{Hexapod -- emerging gaits}\label{sec:hexapod}

\begin{figure}
  \includegraphics[width=\linewidth]{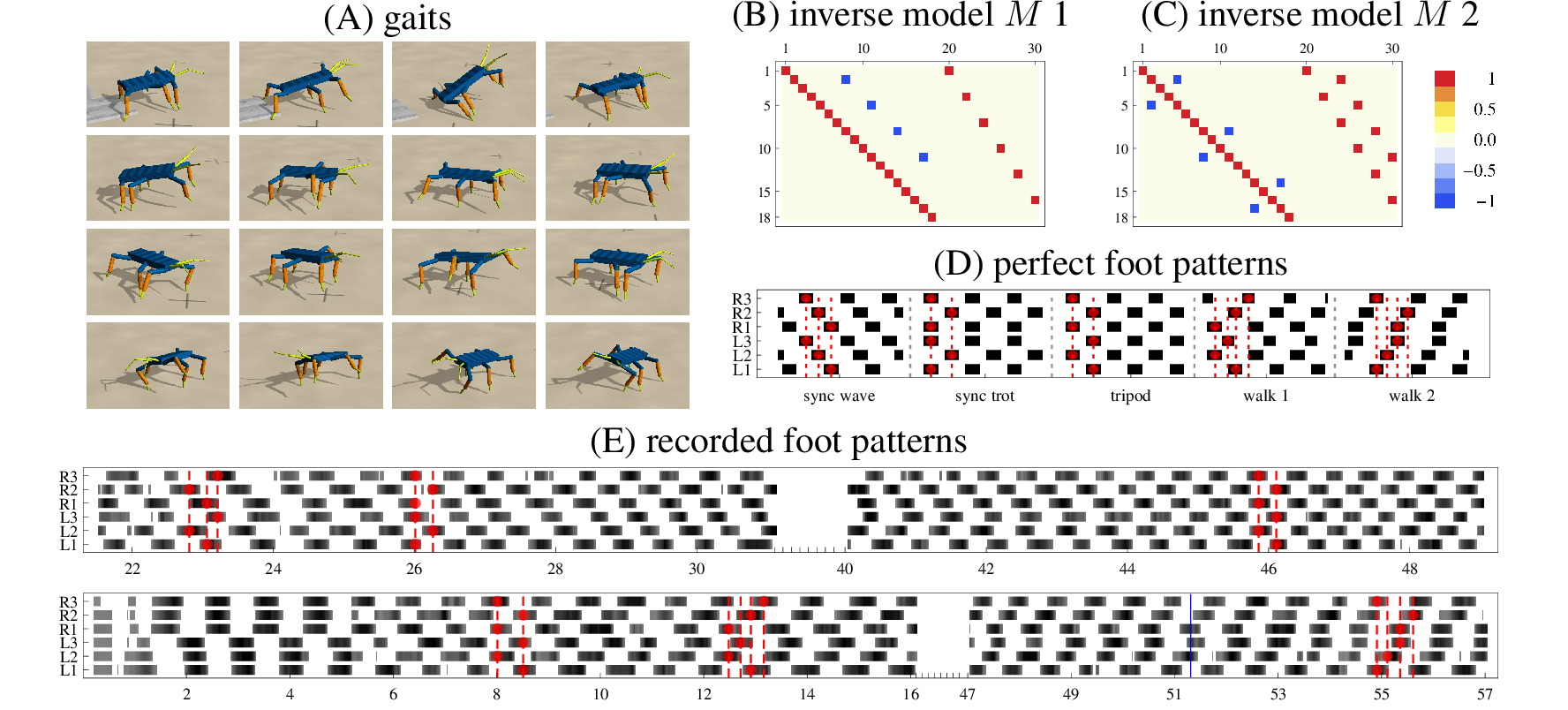}%
  \caption{{\bf Hexapod: emerging gait patterns.}
    The robot is inspired from a stick-insect and has $18$ actuated degrees of freedom (DoF).
    There are $18+12$ sensors (joint angle + delayed ones by $0.2$\,sec).
    The robot performs different gaits when controlled with DEP:
    \emph{synchronous wave} (A row 1), \emph{synchronous trot} (A row 2),
    \emph{tripod} (A row 3),  \emph{walk 1} (A row 4), and \emph{walk 2} with their corresponding step
    pattern in (D) (black means leg is down).
    The fixed inverse models $M$ in two configurations are displayed in (B,C).
    Recorded foot patterns for model $M$ 1 (E top) and model $M$ 2 (E bottom)
    show the transitions between different gaits.
    These transitions are either spontaneous or induced by interactions with the environment
    or by changes in the sensor delay. 
    Markers (red dashed lines and points) indicate the gait patterns for \emph{sync.~wave, sync.~trot, tripod} (top)
    and \emph{tripod, walk 1} and \emph{walk 2} (bottom).
    See also \vid{hexa_loco}.
  }\label{fig:hexa:gaits}
\end{figure}

In the humanoid robot case, the preference of forward locomotion can be related back to the specific
geometry of the body. If on its hands and knees, the lower legs so to say break the forward-backward symmetry
so that backward locomotion is more difficult to achieve.
Let us consider now the hexapod robot, \fig{fig:hexa:gaits},
 that has almost perfect forward-backward symmetry, which must be broken for a locomotion pattern.
This may happen spontaneously but in most experiments motions like swaying or jumping
 on the spot are observed.

Let us now demonstrate how the system can be guided to break its symmetries in a desired way.
This method has essentially two elements.
On the one hand, we have to provide additional sensor information to facilitate circular leg movements.
This is done by providing the delayed sensor values of the 12 \emph{coxa}-joint sensors.
On the other hand,  guidance is implemented by structuring the inverse model $M$ appropriately by hand,
 which is a new technique for guided self-organization of behavior~\cite{martiusherrmann:variantsofgso12}.
The rationale is that those connections in the model are added where correlations
 in the velocities are desired.
For the oscillations of the legs, the delayed forward-backward (anterior/posterior) sensor is linked to the up-down (dorsal/ventral) direction.
In order to excite locomotion behavior,
 the properties of desired gaits can be specified such as in-phase or anti-phase relations of joints.
We present here two possibilities where only few such relations are specified
 in order to give room for multiple behavioral patterns.

In the first configuration the anterior/posterior direction of subsequent legs should be anti-phase resulting in only 4 negative entries in $M$, see \fig{fig:hexa:gaits}(B). Note, there are no connections between the left and right legs.
In the experiment the robot performs the first locomotion pattern already after a few seconds.
A \emph{sync wave} gait emerges where legs on both sides are synchronous
 and hind, middle, and front leg-pairs touch the ground one after another.
This transitions into a \emph{sync trot} gait where hind and front legs are additionally synchronized.
After a perturbation (from getting stuck with the front legs)
 the common \emph{tripod}~\cite{Wilson1966:InsectWalking}(type c) gait emerges, see \fig{fig:hexa:gaits}.

The second configuration resembles what is observed in biological hexapods, namely
 that subsequent legs on each side have a fixed phase shift (which we do by linking
  in $M$ the delayed sensor and motor of subsequent legs) and that legs on opposite sides are anti-phasic.
This results in model $M$ 2, see~\fig{fig:hexa:gaits}(C).
In the experimental run the initial resting state develops smoothly to the \emph{tripod} gait.
Decreasing the time delay of the additional sensors leads first to a gait with seemingly inverse
 stepping order which we entitle \emph{walk 1} and which is also observed in insects~\cite{Wilson1966:InsectWalking}(type f).
For a smaller delay an inverted ripple gait~\cite{Collins1993:hexapodDS}
 appears that we call \emph{walk 2}, see \fig{fig:hexa:gaits} and \vid{hexa_loco}.

An important feature of these closed-loop control networks
 is that they can be used to control a non-trivial behavior with fixed synaptic weights --
 obtained by taking snapshots or from clustering, see \figapp{fig:clusters}.
Behavior sequences can easily be generated by just switching between these fixed sets of synaptic weights.
We demonstrate this with the humanoid robot with different crawling modes in \vid{recall}(A)
 and with the hexapod robot by sequencing all of the emergent gaits in \vid{recall}(B).
Notably, the transition between the motion patterns is smooth and autonomously performed.

\subsection{Finding a task in the world}\label{sec:wheel}
\begin{figure}
  \centering
  \includegraphics[width=\linewidth]{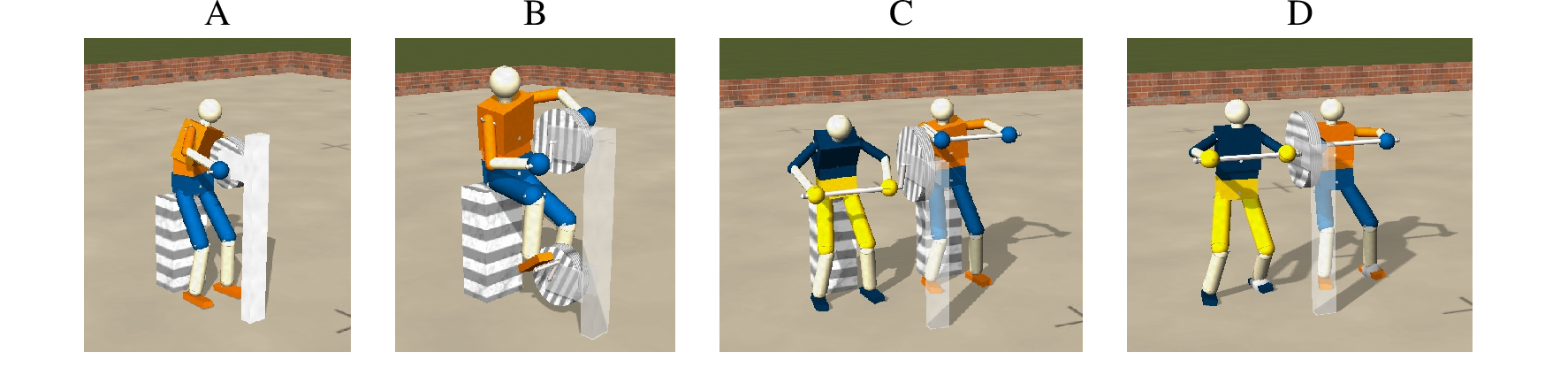}%
  \caption{{\bf Interacting with the environment.}
    The humanoid robot at the wheel with cranks (A), with two wheels (B), and
    with two robots, each at one of the handles, sitting (C) and standing (D).
    See \vid{humanoid_trainer_dhl_noh} and \ref{vid:humanoid_twotrainer_dhl_noh}.
  }
  \label{fig:wheel}
\end{figure}

Up to now, we have seen how the DEP rule bootstraps specific motion patterns contingent on the
physical properties of the body in its interaction with a static environment.
There is a new quality of motion patterns  if the robot is interacting with a dynamical, reactive  environment.
For a demonstration we consider a robot sitting on a stool with its hands attached to the cranks of
a massive wheel, see \fig{fig:wheel}.

\paragraph{Rotating the wheel:}

Different from a static environment, the massive wheel exerts reactive forces on the robot depending on its angular velocity
which is a result of the robot's actions in the recent  past.
 This response and its immediate influence on the learning
process initially leads to a fluctuating relation
between wheel and robot which eventually becomes amplified to end up in a meta-stable periodic motion,
see \vid{humanoid_trainer_dhl_noh}(A).
Interestingly, this effect depends crucially on the mass of the wheel --- it must be large enough so that, by its inertia,
it really can give a feedback to the robotic system.

From the point of view of an external observer, one may say that
the learning system is keen on finding a task in the world
(rotating the wheel) which channels its search into a definite direction.
It cannot be stressed enough that the robot has no knowledge whatsoever about the physical properties
and/or the position and dynamics of the wheel.
All the robot has is the physical answer of the environment (the wheel) by the reactive forces.
In this way the robot is detecting affordances~\cite{Gibson1977:Affordances} of the environment. 
An affordance is the opportunity to perform a certain action with an object, like a chair affords sitting and a wheels affords turning.

The constitutive role of the body-environment coupling is also seen if a torque is applied to the axis of the
wheel. By this external force we may give the robot a hint what to do. When in the fluctuating
phase, the torque immediately starts the rotation which is then taken over by the controller.
Otherwise, we can also ``advise'' the robot to rotate the wheel in the opposite direction, see \vid{humanoid_trainer_dhl_noh}(A).
This can be considered as a kinesthetic training procedure, helping the robot in finding and
realizing its task by direct mechanical influences.

\paragraph{Multitasking}
In a variant of this experiment, also the feet get attached to a separate wheel.
Interestingly, because of the simpler physics, the leg-part of the robot needs much less
time for finding its task as compared to the upper body, see \vid{humanoid_trainer_dhl_noh}(B).
Noteworthy is also the lack of synchronization between the two subsystems. At first sight,
this is no surprise as the upper and the lower body are physically separated completely (the robot is rigidly fixed on the stool).
However, there is an indirect connection given by the fact that each subsystem sees the full set of sensor values.
Actually, this might support synchronizing given the correlation affine properties of the DEP rule.
Yet, due to the largely different physics, synchronization occurs only temporarily, if at all,
so that a kind of multitasking emerges.

\paragraph{Emerging cooperation:}
We have seen above how the exchange of forces with the environment may guide the robot into specific modes.
In this paragraph we show how this can be extended to interacting robots by coupling them  physically or letting them exchange information.
For a demonstration,  we extend the wheel experiment by having two robots,
 each driving one of the cranks, see \fig{fig:wheel}(C-D).
 In this setting, the robots can ``communicate'' with each other by the interaction forces transmitted by the wheel.
 So, by the induced perturbations of its proprioceptive sensor values, each robot can perceive to some degree what the other one
 is doing. By the DEP rule
these extrinsic effects can get amplified, leading eventually to a synchronized motion,  see \vid{humanoid_twotrainer_dhl_noh}.
Seen from outside, the robots must cooperate in order to rotate the wheel
 and it is astonishing that this high-level effect emerges in a natural way from the local plasticity rule.

\section{Discussion}\label{sec:discussion}
This paper reports a simple, local, and biologically plausible synaptic mechanism
 that enables an embodied agent to self-organize its individual sensorimotor development.
 The reported \emph{in silico} experiments have shown  that, at least on the level of sensorimotor
contingencies, self-determined development can be grounded
in this synaptic dynamics, without having to postulate any higher level constructs
such as intrinsic motivation, curiosity, goal orientation, or a specific reward system.
The emerging behaviors --- from various locomotion patterns
 to the ``understanding'' of how to rotate a wheel, to the
spontaneous cooperation induced by force exchange with a partner --- realize
a degree of self-organization unprecedented so far in artificial systems.
These do neither require modifications of the DEP rule nor any task-specific information.
Our demonstrations use artificial systems,
 which is often a problem in computational neuroscience and robotics due to a discrepancy between
 the behavior of artificial and real systems.
In our approach this discrepancy problem is circumvented,
 as the behavior is not the execution of a plan,
 but is emerging from scratch in the dynamical symbiosis of brain, body, and environment.
Commonly, learning to control an actuated system faces the curse of dimensionality, both in model and reality.
Without a proper self-organization process or hand-crafted constraints,
 adding one actuator leads to a multiplicative increase in the time required to find suitable behaviors.
We provide evidence for adequate scaling properties of our approach
 with systems up to $18$ actuators developing, without any prestructuring,
 useful behaviors within minutes of interaction time.
Additionally, we have given arguments on a system theoretical level, that, by being at the edge of chaos
 and allowing for spontaneous symmetry breaking,
 our approach may scale up to systems of biological dimensions,
 like humans with their hundreds of skeletal muscles.

The presence of the DEP rule in nature  may change our understanding of the early stages of sensorimotor development,
 as it introduces an apparent goal orientation and  a self-determined restriction of the search space.
It still remains an open question whether nature found this ``creative'' synaptic dynamics.
The simplicity of the neural control structure and the DEP rule,
 combined with its potential to generate fitness relevant behavior, are strong arguments for
 evolution having discovered it.
In addition the synaptic rule has a simple Hebbian like
structure and may be implemented by combining homosynaptic and heterosynaptic (extrinsic) plasticity mechanisms~\cite{Bailey2000:heterosynaptic} in real neurons.

The DEP rule may also explain leaps in natural evolution.
It is commonly assumed that new traits are the result of a mutation in morphology accompanied by
an appropriate mutation of the nervous system, making the likelihood of selection very low,
 as it is the product of two very small probabilities.
With DEP, new traits would emerge by mutations of the morphology alone.
For instance, the fitness of an animal evolving from water to land
 will be greatly enhanced if it can develop a locomotion pattern on land in its individual life time,
 which could easily be achieved by the DEP rule.
Following the argument by Baldwin~\cite{baldwin1896,weber2003evolution},
 the self-learning process could be replaced in later generations by a
 genetically encoded neuronal structure making the new trait more robust.

Adaptability to major changes in morphology may also be necessary
 for established species during their lifetime.
For instance, injuries such as leg impairments or losses have to be accommodated.
It has long been known that even small animals such as insects have this capability and substantially
 reorganize their gaits patterns~\cite{Wilson1966:InsectWalking}.
This could be achieved with special mechanisms, but with DEP it comes for free.

Another point concerns the role of spontaneity and volition in nature.
Obviously, acting spontaneously is an evolutionary advantage
as it makes prey less predictable to predators.
Attempts to explain spontaneity and volition range
from ignoring it as an illusion to rooting it deep in thermodynamic and even quantum mechanical
randomness~\cite{Brembs2011,Koch2009}.
We cannot give a final  explanation, but the DEP rule provides a clear example how
 a great variety of behaviors can emerge spontaneously in deterministic systems by a deterministic controller.
The new feature is the role of spontaneous symmetry breaking in systems at the edge of chaos.
Similarly, there are recent trends in explaining the apparent stochasticity of the nervous system
 by the complexity of deterministic neural networks~\cite{TrieshEtAl:noise,KenetBibitchkovTsodyks2003:spontaneous,BourdoukanEtAl2012:VariabilityFromOptionalCode}. 

This paper studies a neural control unit in close interaction with the physical environment.
Yet, DEP may also be effective in self-organizing the internal brain dynamics by
considering feedback loops with other brain regions.
This is possible  as the DEP approach does not need an accurate model of the ``rest of the
brain'' --- which could never be realized --- but requires only a coarse idea of the causal features of the response of the system.
In this context our study may provide ingredients required for the big neuroscience initiatives~\cite{KandelEtAl2013:BigNSProjects}
to understand and subsequently realize the functioning brain.

\section{Methods}\label{sec:methods}

\subsection{Normalization} \label{sec:normalization}
For controlling the robot we use
 normalized weight matrices $C$ in \eqn{eqn:DHL10}.
We have the option to perform a \emph{global} normalization or an \emph{individual} normalization for each neuron:

\begin{center}
\begin{tabular}{p{.45\linewidth}|p{.45\linewidth}}
\bf Global normalization & \bf  Individual normalization\\
\hline
the entire weight matrix is normalized&
each motor neuron is normalized individually\\
$C \leftarrow \kappa C/{(\|C\|+\rho)}$&$C_{ij} \leftarrow \kappa C_{ij}/{(\|C_{i}\|+\rho)}$
\end{tabular}
\end{center}

For the global normalization the Frobenius norm $\|C\|$ is used and
for the individual normalization $\|C_{i}\|$ denotes the norm of the $i$-th row (length of synaptic vector of neuron $i$).
The regularization term $\rho = 10^{-12}$ becomes effective near the singularity at $C=0$ or $C_i=0$, respectively,
  and keeps the normalization factor in bounds.
Neurobiologically the normalization can be achieved by a balancing inhibition on a fast time-scale accompanied
 by homeostatic plasticity on a slower time-scale.

With  $\kappa$ small (compared to 1),
activity breaks down so that the system converges towards the resting state where $\dot x=0$.
With  $\kappa$ sufficiently large, this global attractor is destabilized so that modes start to self-amplify
ending up in full chaos for large $\kappa$.
Within an appropriate range of values for $\kappa$,
the system is led towards an exploratory but still controllable behavior.

The type of normalization has a strong impact on the resulting behavior.
The individual normalization leads to behaviors that involve all motors because
 each motor neuron is normalized independently.
To the contrary, with global normalization the overall activity can become restricted to a subset of
 motors.
An example for global normalization is the humanoid robot at the wheel, see \vid{humanoid_trainer_dhl_noh}(A),
 where the legs are inactive because initially the arms are moving more strongly, such that only
 correlations in the velocities of the upper body build up.

\subsection{How it works}\label{sec:suppl:How}

Let us start with a fixed point analysis of the Hebbian learning case with $M$ being the unit matrix
(as in the  humanoid case). Ignoring nonlinearities for this argument (small vectors $x$ and $y$),
the dynamics in the model-world is $x'=y=\alpha Cx$ where $\alpha$ is the normalizing factor.
The system is in a stationary state if $x'=x$ which is the case if $x$ is an  eigenvector of $C$ with eigenvalue $1/\alpha$.
In such a state the learning dynamics ($\tau\dot C_{ij} = y_i x_j-C_{ij}$) converges toward $C=xx^\T$, due to the decay term.
Using this in the stationary state equation ($x=\alpha C x$) yields the condition $x=\alpha x \|x\|^2$ such
 that any vector $x$ with $\|x\|^2 = 1/\alpha$ can be a stationary state, a Hebb state, of the learning system.
This means that the Hebb rule generates a continuum of stationary states
instead of an exploratory behavior.

In the plain DHL case exists a global stationary state if the system is at rest so that $C=0$
(recall plasticity rule: $\tau\dot C_{ij} = \dot y_i \dot x_j-C_{ij}$).
However, if $C \not=0$, the normalization may counteract the effect of the decay term in the plasticity rule.
Thus, Hebb like states can be stabilized for a while until the regularization term $\rho$
 stops this process and $C$ decays to zero.
The thereby induced decay $y\rightarrow 0$ generates motion in the system, leading to a learning signal.
With $\kappa$ large, this may avoid the convergence altogether so that the system
can find another Hebb state and so forth.
However, the activities elicited by this interplay between the normalization and the learning dynamics
are rather artificial and less rich as they do not incorporate the extrinsic signals given by $\delta y$.
In particular, the plain DHL rule has no means of incorporating the extrinsic effects provided
 by additional sensor information as described in \sec{sec:lr:comparison}.

The novel feature of the proposed DEP rule \eqn{eqn:DHL100simple}
 is the leading role of the extrinsic signal ($\delta y$).
For a demonstration, consider the trivial case  $\dot y=0$ where the body is at rest and will stay there as long as
 there are no extrinsic perturbations. However,
if the body is being kicked by some external force,  $x$, $x'$ and hence $\tilde{\dot y}$ may vary
so that $C$ changes and the system is driven out of the global attractor if $\kappa$ is sufficiently large.
In fact we observe in the experiments how an initial kick  acts like a dynamical germ for the starting of
an individual behavior development. Moreover, also in the behaving system, external influences may
change the behavior of the system, see for instance
the gait switching of the hexapod after the perturbation by the obstacles (\vid{hexa_loco})  or
 the phenomenon of emerging cooperation of the humanoids (\vid{humanoid_trainer_dhl_noh}).
A quantitative comparison of the learning rules is given in \sec{sec:lr:comparison}.

The next remark concerns the dichotomy between the learning as formulated in the velocities and the
control of the system based on the (angular) positions, which is vital for the ``creativity'' of the DEP rule.
This feature of the DEP rule \eqnp{eqn:DHL100simple} can be elucidated by a self-consistency argument,
assuming the idealized case the system
is in a harmonic oscillation (an idealized walking pattern, \eg) with period $T$.
By way of example, let us consider the two-dimensional case with a rotation matrix $U(s)$,
generating a periodic motion by
rotating a vector $x(\alpha)=(\cos \alpha,\sin \alpha)^\T$ by an angle $s$, \ie $U(s)x(\alpha)=x(\alpha+s)$.
We use that $x'$ is obtained by the time evolution of $x$ over a finite time step so that $x'(\alpha)=x(\alpha + \theta)=U(\theta )x(\alpha)$
and $\langle \dot {x}^\prime \dot x^\T \rangle = U(\theta ) \langle \dot {x} \dot x^\T \rangle$.
Taking the average over one period we have $\langle \dot {x} \dot x^\T \rangle \propto \langle {x} x^\T \rangle $
as $\dot x$ is a phase shifted copy of $x$ times a factor.
For the averaging, we consider infinitesimally short time steps so that sums may be replaced with integrals.
We obtain $\langle \dot {x} \dot x^\T \rangle\propto
\int x(\alpha) x(\alpha)^\T d\alpha\propto I$, where $I$ is the unit matrix, because
$\int \cos^2(\alpha)d\alpha= \int \sin^2(\alpha) d\alpha\propto 1$ and  $\int \sin(\alpha)\cos (\alpha)d\alpha=0$.
Neglecting the extrinsic effects, \ie $\delta y=0$, under these idealized conditions, we
may write in the linear regime $Mx'=y=Cx$ and with
$C=M\langle \dot {x}^\prime \dot x^\T \rangle\propto MU(\theta)$
we  obtain  $x'=U(\theta )x$ corroborating that the periodic regime is stationary if the
gain factor is chosen appropriately.

In higher dimensions the argument is more involved but it carries over immediately to the case of rotations about an axis,
corresponding to the case that the Jacobian matrix $J=MC$ has
just one pair of complex eigenvalues different from zero. Interestingly, this situation is
often observed in the experiments, with the hexapod, \eg, even though these experiments are very far
from the idealized conditions postulated for the above proof.

These patterns are self-consistent solutions of the mentioned  dichotomy between control itself and the learning
generating it.
As the experiments show, these patterns are metastable attractors of the whole-system dynamics, so that globally,
the system realizes a ``search and converge'' strategy
by switching between metastable attractors via possibly long transients.
At the behavioral level,
this is reflected by the emergence of a great variety of spatio-temporal patterns---the
global order obtained from the simple local rules given by \eqns{eqn:DHL100simple}{eqn:biasdyn}.

\subsection{Comparison of synaptic rules}\label{sec:lr:comparison}
\begin{figure}
  \centering
  \includegraphics[width=\linewidth]{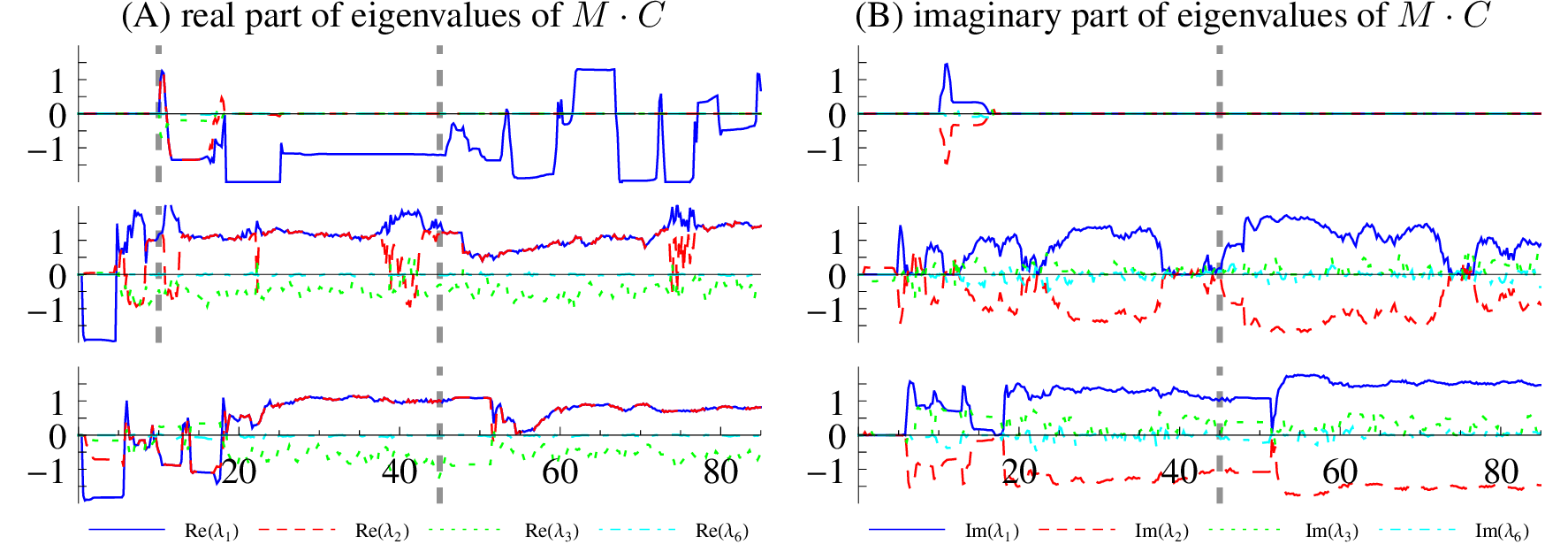}%
  \caption{{\bf Comparison of plasticity rules: DHL, DEP, and DEP-Guided (with structured model) (from top to bottom).}
    Development of the eigenvalues $(\lambda_1,\lambda_2,\lambda_3,\lambda_{6})$, real part in~(A) and imaginary in~(B),
    of the matrix $M\cdot C$ over time.
    After 10\,sec  (dashed vertical line) the synaptic weights from DEP were copied to DHL
     in order to get some non-zero initialization.
    The dashed vertical line at 45\,sec marks a strong physical perturbation which leads to changes in behavior
    for the DEP rule, but not for DHL. See \vid{hexa_comparison}.
    Only for DEP the eigenvectors (see \vid{hexa_comparison}) and the spectrum changes significantly after the perturbations.
    Parameters: global normalization, $\kappa=2.2, h=0, \tau=0.7$\,sec.
  }
  \label{fig:lr:comparison}
\end{figure}

In this section we want to compare the different synaptic rules.
We will consider them in their pure form without the threshold dynamics, \ie $h=0$.
Also in experiments we can confirm that plain Hebbian learning produces fixed point behaviors
 and thus no continuous motion.
Therefore we will only compare DHL with DEP at the example of the hexapod robot.
In order to provide a fair comparison the hexapod robot is started in three identical experiments,
except for the plasticity rules:
DHL: $\tau\dot C_{ij} =  \dot y_i \dot x_j -  C_{ij}$,
DEP:  $\tau\dot C_{ij}= \tilde{\dot y_i } \dot x_j -  C_{ij}$ \eqnp{eqn:DHL100simple} (with unit model),
and DEP-Guided with model $M$ 1, see (\fig{fig:hexa:gaits}(B)).
Note, that the same synaptic normalization and decay is used everywhere.
Since DHL is \emph{not} able to depart from the $C=0$ condition,
 we copy the synaptic weights ($C$) of the DEP run to the DHL experiment after $10$\,sec.
We then consider the eigenvalue spectrum and the corresponding eigenvectors of the matrix $R=M\cdot C$.
(for DHL and DEP $M=\mathbb I$).
The matrix $R$ captures the linearized mapping from $x$ to $x'$ -- the dynamics of the sensors.
For DHL the matrix reduces to have only a single non-zero eigenvalue. This in turn means that
 all future sensor values are projected onto the corresponding eigenvector
 and the learning dynamics cannot depart from that. This is demonstrated in
\figapp{fig:lr:comparison} and \vid{hexa_comparison}.
Under heavy perturbations the robots controlled by DEP change their internal structure and
 show subsequently a different behavior.
For a different initialization DHL (with normalization) may also produce continuous motion patterns
 but it is generally much less sensitive to the embodiment and perturbations
 and often falls into the $C=0$ state, which it cannot exit anymore.

\subsection{Framework}
The use of artificial creatures has proven
a viable method for testing neuroscience hypotheses,
providing fresh insights into the function of the nervous system, see~\cite{Floreano2014:RoboticsAndNeuroscience}, \eg 
The benefit of such methods may be debatable with brain processes of higher complexity.
So, we focus on the lower level sensorimotor contingencies  in settings where
the nervous systems
 cannot be understood in isolation due to a strong brain-body-environment
 coupling~\cite{ChielBeer1997:behaviorfrominteraction,PfeiferBongar2006:BodyShapesThink}.
The method works both with hardware robots or with good physical simulations.
The inexorable reality gap between simulation and real robotic experiments
 is a minor methodological problem because the phenomenon of self-organization
 is largely independent of the particular implementation, as evidenced by
 the fact that emerging motion patterns are robust against modifications of the
 physical parameters of both robot and environment.

The control structure in this paper is deliberately chosen simple in order to demonstrate the potential of the method.
Although in larger animals the time lag of sensory feedback is too long for this setting,
 it may still be successful by integrating sensor predictions, a trick the
nervous system is using whenever possible~\cite{Massion1994:Posturalcontrol}.
In the following we discuss several of the methodological issues in detail.

\paragraph{Initialization:}
In order to have reproducible conditions for our experiments, we always start our system
under definite initial conditions.
We use in all our applications the same plasticity rule
(with appropriately chosen time scale $\tau$ and gain factor $\kappa$) and start moreover
always in the same initial conditions for the synaptic dynamics by choosing $C=0$ and $h=0$, meaning that
all actuators of the joints are in their central  position.
In a sense, this is also a state of
maximum symmetry, as there is (approximately) no difference in moving the joint either forward or backward,
so that the observed behaviors are emerging by spontaneous symmetry breaking~\cite{DerGSO2012,DerMartius13}.

\paragraph{Source of variety: }
In common robotics approaches to the self-exploration of sensorimotor contingencies,
innovation is introduced by modifying actions randomly~\cite{Schmidhuber91:CuriousControl,oudeyer07:IntrinsicMotivation},
relegating the activities of the agent to the intrinsic laws of a pseudo-random number generator
that is entirely external to the system to be controlled.
This  invokes the curse of dimension as the search is not restricted by the
specifics of the brain-body-environment coupling.
In our system, there is only one intrinsic mechanism -- the DEP rule --
which generates actions deterministically in terms of the sensor values over the recent past
(on a time scale given by $\tau$). Variety is produced by the complexity of the physical world in the sense of deterministic chaos.
Another approach to obtain informed exploration is to employ notions of information gain~\cite{FrankSchmidhuber2013:curiosity,LittleSommer2013:PIG}
 which are unfortunately so far not scalable to the high-dimensional continuous systems used here.

\paragraph{Simulation:}
The experiments are conducted in a physically realistic rigid body simulation tool \textsc{Lpz\-Robots}~\cite{lpzrobots10}. 
The tool is open source so that the experiments of this paper can be reproduced, see \app{sec:supp:source}.
The humanoid robot, \fig{fig:motorbabbling},  has the proportions and weight distributions of the human body.
The joints are simplified and only one-axis and two-axis joints are used.
The degrees of freedom (DoF) are: 4 per leg (2 hip, knee, ankle), 3 per arm (2 shoulder, elbow), 1 for the pelvis (tilting the hip) and 3 in the back (torsion, bending front/back and left/right) summing up to 18 DoF.

The hexapod robot, \fig{fig:hexa:gaits}, is inspired from a stick-insect and has 18 DoF, 3 in each leg: two in the coxa joint and one in the femur-tebia joint ($\alpha,\beta,\gamma$ in \cite{SchillingCruse2013:Walknet}). The antennas and tarsi are attached by spring joints and are not actuated.

To implement the actuators we use position controlled angular motors with strong power constraints
 around the set point to make small perturbations perceivable in the joint-position sensors.
In this way they perform more like muscles and tendon systems
 and make the robot compliant to external forces.

\paragraph{Internal model:}

The appropriate choice and internal representation of inverse models, projecting the sensor values back to the causing motor commands,
are topics of much interest in both robotics and neuroscience. Modeling this relationship
is generally thought to be a very complex task.  As our
experiments have shown, different from the realization of a specific behavior (like reaching or grasping),
the exploration, which has no predefined goal, can work with crude models
(like the matrix $M$ in \eqn{eqn:hypotheticalcommand}),
producing complex motion patterns which never could be adequately represented by the model.
While the specific route followed by the self-exploration
may well depend on the quality of the model,
self-organization itself is a robust phenomenon that prevails
also under obstructive circumstances. The deeper reason is found by understanding behavior as broken symmetry,
and individual development as a series of spontaneous symmetry breaking events. 
By their very nature, the latter are robust phenomena rooted in
the system dynamics,  but their sequence is prone to the actual situations---in our case the projections made by the model.

\subsection*{Acknowledgments}
The authors gratefully acknowledge the hospitality and discussions in the group of Nihat Ay.
We thank Michael Herrmann and Anna Levina and Sacha Sokoloski for helpful comments.
GM was supported by a grant of the DFG (SPP 1527) and received funding from the People Programme (Marie Curie Actions) of the European Union's Seventh Framework Programme (FP7/2007-2013) under REA grant agreement no.~[291734].


\appendix
\newcommand{\latex}[1]{#1}
\newcommand{\videopage}[1]{}


\latex{
  \newcommand{\videoname}{Video}
  \newcounter{video}
  \renewcommand{\thevideo}{\arabic{video}}

  \newcommand{\Video}[5][.4\linewidth]{
    \nobreak%
    \begin{center}
      \href{http://playfulmachines.com/DEP/index.html\#vid:#5}{%
        \includegraphics[width=#1]{#2}}%
    \end{center}\vspace*{-.5em}\nobreak%
    Video \refstepcounter{video}\thevideo: #3 #4
      See {\footnotesize\href{http://playfulmachines.com/DEP/index.html\#vid:#5}%
        {\nolinkurl{http://playfulmachines.com/DEP}.}}%
      \label{vid:#5}%
  }
  \newcommand{\Videowide}[5][.9\linewidth]{\Video[#1]{#2}{#3}{#4}{#5}}
  \newcommand{\Videos}[6][.3\linewidth]{
   \begin{center}
     \begin{tabular}{c@{\qquad}c}
       (A) & (B)\\
       \href{http://playfulmachines.com/DEP/index.html\#vid:#6:a}{
         \includegraphics[height=#1]{#2}} &
       \href{http://playfulmachines.com/DEP/index.html\#vid:#6:b}{
         \includegraphics[height=#1]{#3}}
     \end{tabular}
   \end{center}\vspace*{-.5em}
   Video \refstepcounter{video}\thevideo: #4 #5
   See {\footnotesize\href{http://playfulmachines.com/DEP/index.html\#vid:#6}%
     {\nolinkurl{http://playfulmachines.com/DEP}.}}%
   \label{vid:#6}%
 }

}



\section{Supplementary videos}\label{sec:supp:Overview}
Videos of the conducted experiments are presented in the following sections.
Information on the source-code is given in \app{sec:supp:source}.

\Video[.4\linewidth]{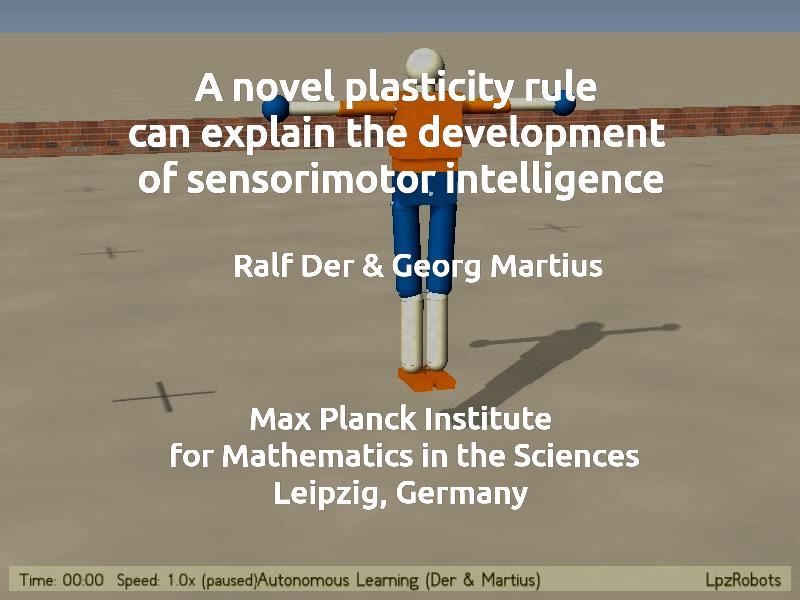}
{Overview video summarizing the experimental results of the paper. }
{This video provides a demonstration of the self-organization of behavior
  created by the novel plasticity rule.
  Different systems are considered and the plasticity rule is briefly explained.
  Longer videos for each experiment are provided below.
}
{DEP-demo} 

\subsection{Self-organized crawling behavior}\label{sec:supp:crawl}
\Video[.4\linewidth]{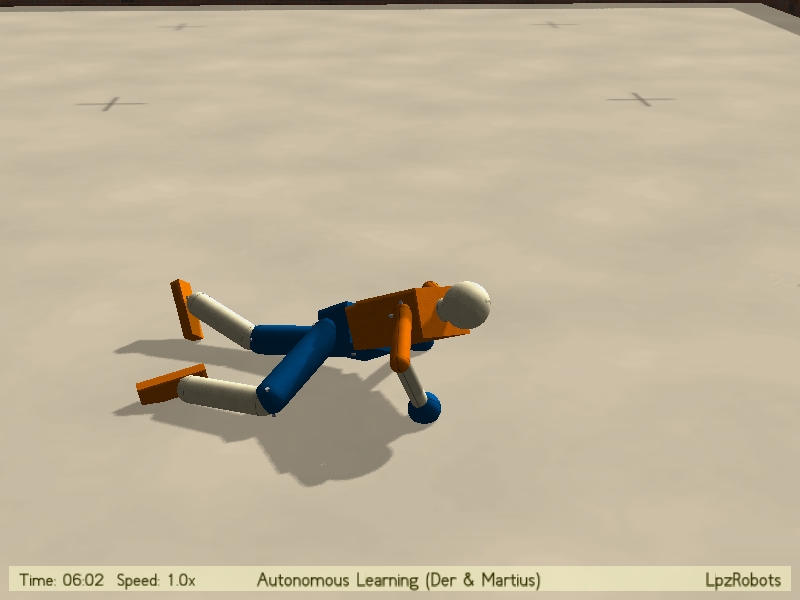}
{The humanoid robot on the floor developing a crawling behavior. }
{During the initial period it is seen how, from the initialization condition, small
 movements get amplified into a coherent movement.
These get increasingly shaped to fit the situation. From time $4$:$30$ on a stable crawling motions is observed.
Parameters: global normalization, $\kappa=1.4, \tau_h=0.4$\,sec$, \tau=4$\,sec.
}
{humcrawl} 

\subsection{Hexapod walking}
\Videos[.25\linewidth]{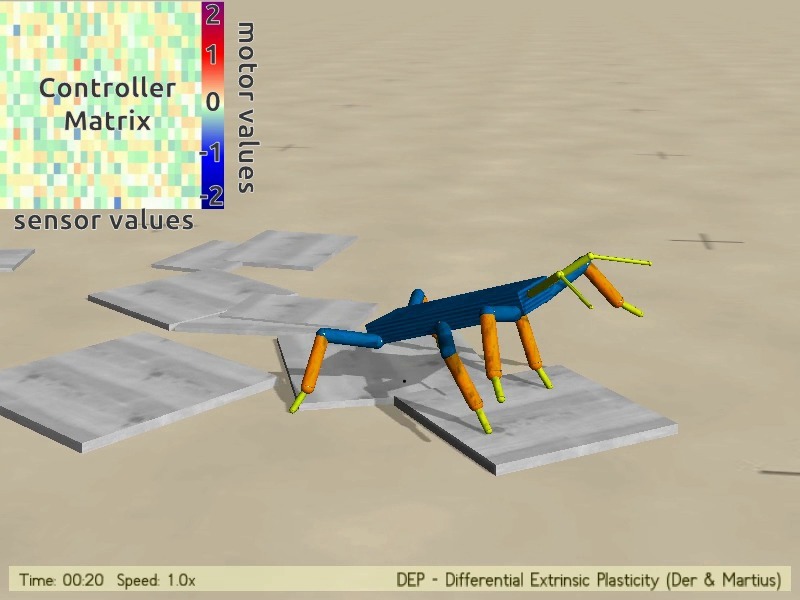}{hexa_loco_delay15-9_lateral-cut}
{Hexapod developing different gaits.}
{ The internal model is structured to break the symmetries and facilitate a locomotion
  behavior, model $M$ 1 (A) and $M$ 2 in (B), see \fig{fig:hexa:gaits}(B,C).
  (A)~Initially a synchronous wave gait and then a synchronous trot gait
  emerges which preserves most of the initial symmetries.
  After a perturbation by getting stuck on the front legs, new locomotion gaits develop which
  channel into the well known tripod gait.
  (B)~Due to a different correlation structure imposed by model $M$ 2,
  a different set of gaits emerges.
  See \fig{fig:hexa:gaits}(D,E) for the stepping patterns.
  Parameters: individual normalization, $\kappa=2.2, h=0, \tau=0.4$\,sec. 
}
{hexa_loco} 

\subsection{Memorizing and recalling behavior}
\begin{figure}[htb!]
  \renewcommand{\tabcolsep}{.1cm}
  \centering
  \begin{tabular}{ccccc}
    \small 1: at the spot& \small 2: high crawling& \small 3: fast crawling&\small 4: bottom sideways&\small 5: forward/backward\\
    \includegraphics[height=.17\linewidth]{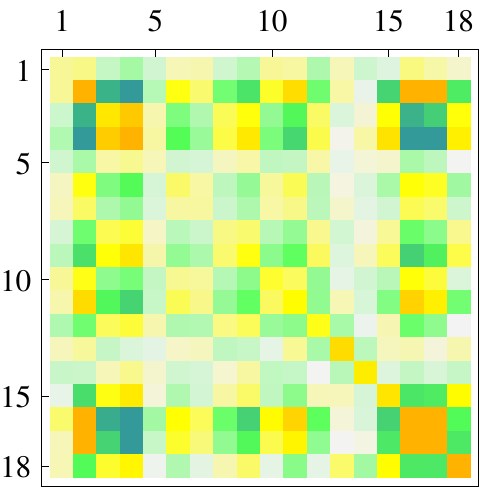}&
    \includegraphics[height=.17\linewidth]{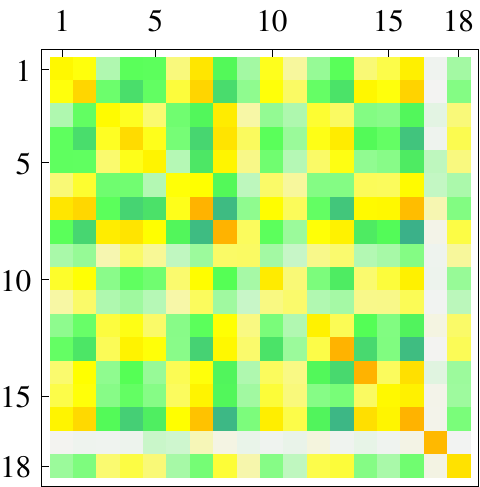}&
    \includegraphics[height=.17\linewidth]{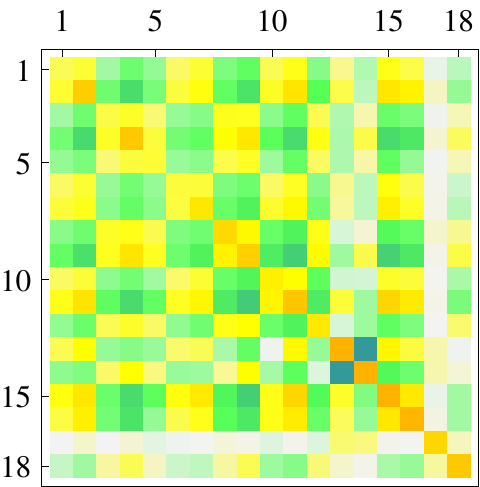}&
    \includegraphics[height=.17\linewidth]{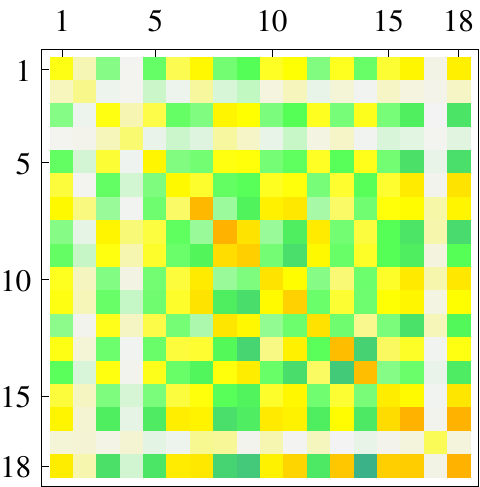}&
    \includegraphics[height=.17\linewidth]{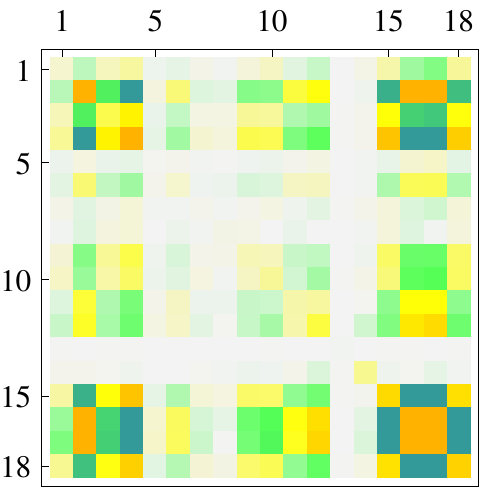}
  \end{tabular}
  \caption{Clustering of the controller matrices of \vid{humcrawl}.
    Displayed are the cluster centers for some of the clusters. The  names were given after a visual inspection
    of their corresponding behavior, see \vid{recall}.}
  \label{fig:clusters}
\end{figure}

By either taking snapshots of the synaptic weights, or more objectively,
 by a clustering procedure, a set of fixed control structures can be
 extracted. Examples of such a clustering are shown in \figapp{fig:clusters}.
If these synaptic connections are used in the controller network,
the behavior can be reproduced (without synaptic dynamics) in most cases.
A demonstration is given in \vid{recall}(A)
 where 5 clusters from \vid{humcrawl}(A,B) have been selected for demonstration.
The switching between behaviors may not be successful if the old behavior does not lie in the basin
 of attraction of the new one. This happens in particular when starting from inactive behaviors, but his can be
helped by  a short perturbation.
In a second experiment we show how snapshots of the weights taken during \vid{hexa_loco}(A,B)
 can be used to memorize and recall all the different emergent gaits.
The synaptic weights have been copied instantaneously without selecting a
 precise time point or averaging.
In fact, during learning for one particular gait a whole series of weights occurred,
 but apparently any of these weight sets is a viable controller.

\Videos[.25\linewidth]{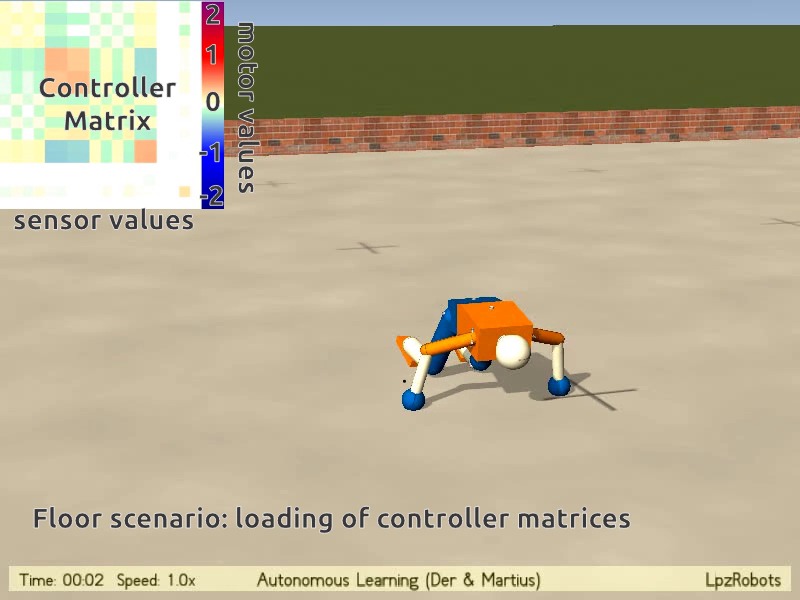}{hexa_control_from_snapshots-cut}
{Recall and transition between behaviors.}
{(A)~Sequence of behaviors generated by switching the synaptic weights to those determined by cluster analysis, see \figapp{fig:clusters}.
Typically, a fast transition between different behaviors occurs. In some cases, however, an external perturbation is required to facilitate the transition.
(B)~For the hexapod robot, a set of matrices have been stored by taking snapshots of the synaptic weights
 as indicated in \fig{fig:hexa:gaits}(E), \ie at seconds: $23, 26, 39$ (faster sync trot, not in \fig{fig:hexa:gaits})$, 46$ of run 1 and at sec $8, 13, 55$ of run 2. No averaging was performed. All previously observed gaits are successfully reproduced with remarkably smooth transition between them.
}{recall} 

\subsection{Humanoid robot at the wheel: find your task in the world}
\Videos[.25\linewidth] {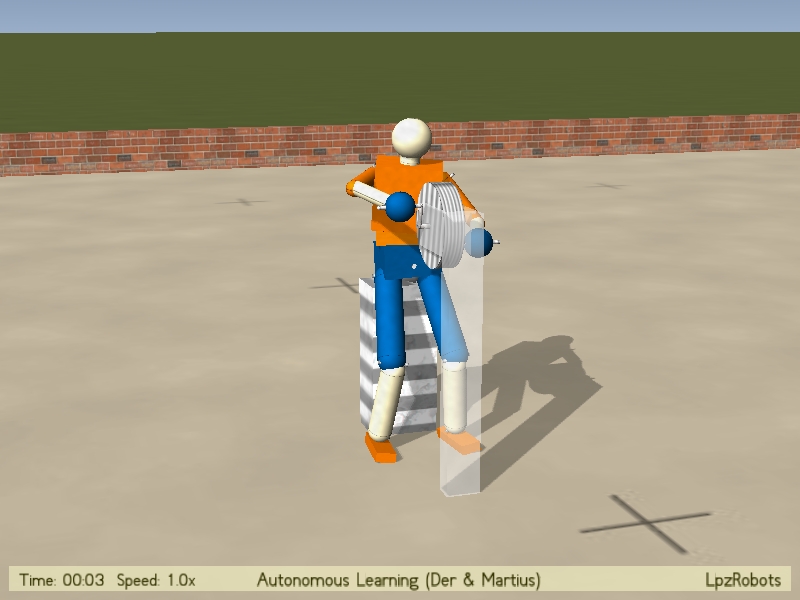}{humanoid_trainer_with_feet_flexible-cut}
{The humanoid robot at the wheels.}
{(A)~The hand of the robot are connected to the cranks of a massive wheel.
  By the drive to build up correlations between joint angle velocities,
  the robot ``discovers'' how the crank can be moved
  in order to realize a stable periodic motion of its internal joint angles.
  When the wheel is turned in the opposite direction by an external torque the
  control network is taking up the new direction quickly.
  Parameters: global normalization, $\kappa=0.96, h=0, \tau=0.4$\,sec. 
(B)~Hand and feet are connected to independent wheels. The feet start to rotate the lower wheel
 earlier than the hands do, due to simpler physics. This difference also hinders spontaneous synchronization.
 Note that the trunk is fixed on the stool such that upper and lower body are physically decoupled.
 Nevertheless, when the lower wheel's revolution direction gets inverted by external force
 the arms also stop and need some time to find the rotating motion again.
 Parameters: individual normalization, $\kappa=1.4, h=0, \tau=0.4$\,sec. 
}
{humanoid_trainer_dhl_noh} 

\subsection{Emerging cooperation}
\Videos[.25\linewidth] {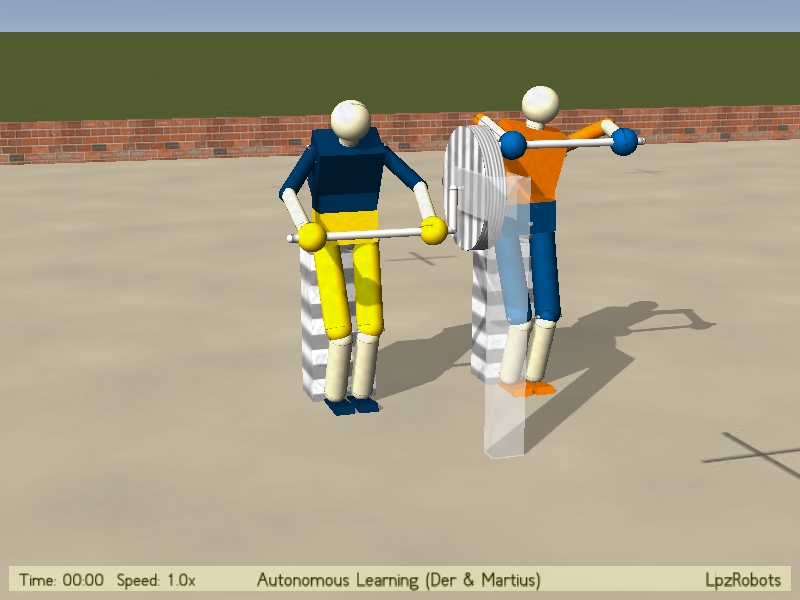}{twotrainerstanding}
{Emerging communications by force exchange.}
{Both robots get connected to the wheel, each to one of the cranks.
The robots have no information about their partner. Yet they manage to cooperate by ``feeling''
 the others reactive forces.
This even works if the robots are not supported by the stool (B) (muscle forces doubled).
Parameters: see \vid{humanoid_trainer_dhl_noh}(A) but with $\kappa=1$.
}
{humanoid_twotrainer_dhl_noh} 

\subsection{Comparison of synaptic rules}\label{app:lr:comparison}
\Videowide[0.7\linewidth]{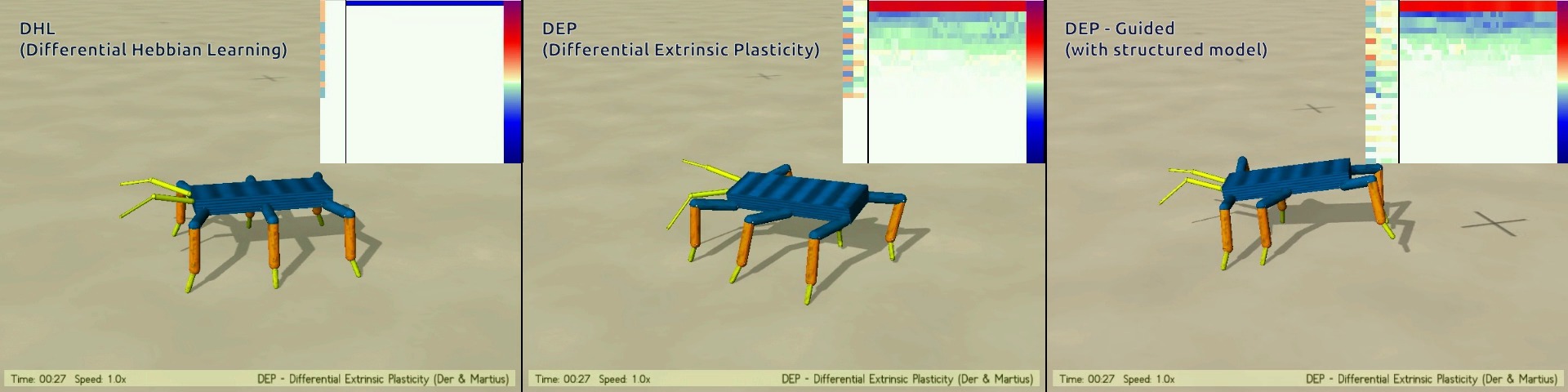}
{Comparison of plasticity rules with the hexapod robot.}
{Left: DHL, middle: DEP, right: DEP-Guided (with structured model, \fig{fig:hexa:gaits}(B)).
The insets show the eigenvalue spectrum over time and
 vertically the first 3 eigenvectors scaled with their corresponding eigenvalues.
The robot is initialized with a zero controller as usual. Only DEP departs from this situtation.
After $10$\,sec the synaptic weights ($C$) are copied from DEP to DHL.
DHL is not able to maintain a persistent motion, even after a perturbation (at $45$\,sec).
Both DEP settings show smooth behavior and a reaction to the perturbation
 by a different motion (middle) or changing the gait pattern (right).
The eigenvalue spectrum differs across the examples.
In the DHL case, there is only one nonzero eigenvalue whereas DEP has about five (see \fig{fig:lr:comparison}). Its effect becomes visible when the robot
 is disturbed after $45$\,sec: DHL remains largely uninfluenced and also the eigenvector
 does not change its direction (it changes intensity in the visualization because it is scaled with the corresponding eigenvalue).
Parameters: see \fig{fig:lr:comparison}.
}
{hexa_comparison} 

\section{Simulation source code}\label{sec:supp:source}
The experiments can be reproduced by using our simulation software and the following sources.
The simulation software can be downloaded from \href{http://robot.informatik.uni-leipzig.de/software}{http://robot.informatik.uni-leipzig.de/software} or
 \href{https://github.com/georgmartius/lpzrobots}{https://github. com/georgmartius/lpzrobots} and has to be compiled on a Linux platform (for some platforms packages are available).
The source code for the experiments of this paper
\latex{\href{http://playfulmachines.com/DEP}{can be downloaded from http://playfulmachines. com/DEP}.}
\videopage{is in preparation.}
Instructions for compilation and execution are included in the bundle.

\end{document}